\documentclass[conference]{IEEEtran}
\IEEEoverridecommandlockouts
\usepackage{cite}
\usepackage{comment}
\usepackage{amsmath,amssymb,amsfonts}
\usepackage{algorithmic}
\usepackage{graphicx}
\usepackage{textcomp}
\usepackage{xcolor}
\makeatletter
\newcommand{\ALOOP}[1]{\ALC@it\algorithmicloop\ #1%
  \begin{ALC@loop}}
\newcommand{\ENDALOOP}{\end{ALC@loop}\ALC@it\algorithmicendloop}

\makeatother
\usepackage[linesnumbered,ruled,vlined]{algorithm2e}
\newcommand{\floor}[1]{\left\lfloor #1 \right\rfloor}
\usepackage[a4paper, total={184mm,239mm}]{geometry}
\def\BibTeX{{\rm B\kern-.05em{\sc i\kern-.025em b}\kern-.08em
    T\kern-.1667em\lower.7ex\hbox{E}\kern-.125emX}}
\begin{document}
\SetKwComment{Comment}{$\triangleright$\ }{}
\renewcommand\algorithmiccomment[1]{%
  \hfill\#\ \eqparbox{COMMENT}{#1}%
}


\title{Can Deep Neural Networks be Converted to Ultra Low-Latency Spiking Neural Networks?}

\author{\IEEEauthorblockN{Gourav Datta and Peter A. Beerel}
\thanks{$^{\dagger}$This work was supported by the NSF CCF-1763747 award.}
\it{Ming Hsieh Dept. of Electrical and Computer Engineering, University of Southern California, Los Angeles, USA} \\
 	\it{\{gdatta, pabeerel\}@usc.edu} \\[-1.0ex]}
\maketitle

\begin{abstract}
Spiking neural networks (SNNs), that operate via binary spikes distributed over time, have emerged as a promising energy efficient ML paradigm for resource-constrained devices. However, the current state-of-the-art (SOTA) SNNs require multiple time steps for acceptable inference accuracy, increasing spiking activity and, consequently, energy consumption.
SOTA training strategies for SNNs involve conversion from a non-spiking deep neural network (DNN).
In this paper, we determine that SOTA conversion strategies cannot yield ultra low latency because they incorrectly assume that the DNN and SNN pre-activation values are uniformly distributed. We propose a new training algorithm that accurately captures these distributions, minimizing the error between the DNN and converted SNN.
The resulting SNNs have ultra low latency and high activation sparsity, yielding significant improvements in compute efficiency. In particular, we evaluate our framework on image recognition tasks from CIFAR-$10$ and CIFAR-$100$ 
datasets on several VGG and ResNet architectures. We obtain top-$1$ accuracy of $64.19$\% with only $2$ time steps on the CIFAR-$100$ dataset with ${\sim}159.2\times$ lower compute energy compared to an iso-architecture standard DNN. Compared to other SOTA SNN models, our models perform inference $2.5$-$8\times$ faster (i.e., with fewer time steps).  
\end{abstract}

\begin{IEEEkeywords}
SNN, DNN, neuromorphic, FLOPs, surrogate gradient learning
\end{IEEEkeywords}

\section{Introduction}\label{sec:introduction}
Spiking Neural Networks (SNNs) 
attempt to emulate the remarkable energy efficiency of the brain in vision, perception, and cognition-related tasks using event-driven neuromorphic hardware \cite{neuro_frontiers}. Neurons in an SNN exchange information via discrete binary spikes, representing a significant paradigm shift from high-precision, continuous-valued deep neural networks (DNN) \cite{spike_ratecoding, dsnn_conversion1}. Due to its high activation sparsity and use of accumulates (AC) instead of expensive multiply-and-accumulates (MAC), SNNs have emerged as a promising low-power alternative to DNNs whose hardware implementations are typically associated with high compute and memory costs.

Because SNNs receive and transmit information via spikes, analog inputs have to be encoded with a sequence of spikes. There have been multiple encoding methods proposed, such as rate coding \cite{diehl2016conversion}, temporal coding \cite{comsa_2020}, rank-order coding \cite{Kheradpisheh_2020}, and others. However, recent works \cite{rathi2020dietsnn, wu2019direct,Kundu_2021_ICCV} showed that, instead of converting the image pixel values into spike trains, directly feeding the analog pixel values in the first convolutional layer, and thereby, emitting spikes 
only in the subsequent layers, can reduce the number of time steps needed to achieve SOTA accuracy by an order of magnitude. Although the first layer now requires MACs, as opposed to the cheaper ACs in the remaining layers, the overhead is negligible for deep convolutional architectures. Hence, we adopt this technique, termed \emph{direct encoding}, in this work. 

In addition to accommodating various forms of encoding inputs, supervised learning algorithms for SNNs have overcome various roadblocks associated with the discontinuous derivative of the spike activation function \cite{lee_dsnn, kim_2020}. Moreover, SNNs can be converted from DNNs with low error by approximating the activation value of ReLU neurons with the firing rate of spiking neurons \cite{dsnn_conversion_abhronilfin}. SNNs trained using DNN-to-SNN conversion, coupled with supervised training, have been able to perform similar to SOTA DNNs in terms of test accuracy in traditional image recognition tasks \cite{rathi2020dietsnn, rathi2020iclr}. However, the training effort still remains high, because SNNs need multiple time steps (at least $5$ with direct encoding \cite{rathi2020dietsnn}) to process an input, and hence, the backpropagation step requires the gradients of the unrolled SNN to be integrated over all these time steps, which significantly increases the memory cost \cite{panda_res}. Moreover, the multiple forward passes result in an increased number of spikes, which degrade the SNN's energy efficiency, both during training and inference, and possibly offset the compute advantage of the ACs. This motivates our exploration of novel training algorithms to reduce both the test error of a DNN and the conversion error to a SNN, while keeping the number of time steps extremely small during both training and inference.

In summary, the current challenges in SNNs are multiple time steps, large spiking activity, and high training effort, both in terms of compute and memory. To address these challenges, this paper makes the following contributions.
\begin{itemize}
    \item We analytically and empirically show that the primary source of error in current DNN-to-SNN conversion strategies \cite{deng2021optimal, li2021free} is the incorrect and simplistic model of the distributions of DNN and SNN activations. 
    \item We propose a novel DNN-to-SNN conversion and fine-tuning algorithm that reduces the conversion error for ultra low latencies by accurately capturing these distributions and thus minimizing the difference between SNN and DNN activation functions. 
    \item We demonstrate the latency-accuracy trade-off benefits of our proposed framework through extensive experiments with both VGG \cite{simonyan2014very} and ResNet \cite{he2016deep} variants of deep SNN models on CIFAR-$10$ and CIFAR-$100$ \cite{krizhevsky2009learning}. 
    We benchmark and compare the models' training time, memory requirements, and inference energy efficiency in both GPU and neuromorphic hardware with two SOTA low-latency SNNs.\footnote{We use VGG$16$ on CIFAR-$10$ and CIFAR-$100$ to show compute efficiency.}
\end{itemize}

The remainder of this paper is organized as follows. Section \ref{sec:background} provides  background on DNNs and SNNs and the SOTA DNN-to-SNN conversion techniques. Section \ref{sec:ann_snn_conversion_failure} explains why these techniques fail for ultra-low SNN latencies and discusses our proposed methodology. Our accuracy and latency results are presented in Section \ref{sec:expt_results} and our analysis of training resources and inference energy efficiency is presented in Sections \ref{sec:train_memory_analysis} and  \ref{sec:energy_analysis}, respectively. The paper concludes in Section \ref{sec:conclude}.
\vspace{-1mm}
\section{Background}\label{sec:background}
\subsection{Difference between DNNs and SNNs}\label{sec:background}
Neurons in a non-spiking DNN integrate weight-modulated analog inputs and apply a non-linear activation function. Although ReLU is widely used as the activation function, previous work \cite{ho2021tcl} has proposed a trainable threshold term, $\mu$, for similarity with SNNs. In particular, the neuron outputs with threshold ReLU can be expressed as
\begin{align}
Y_i &= \text{clip}\left(\sum_{j}W_{ij}X_{j},0,\mu\right)
\label{eq:dnn_activation}
\end{align}
where $\text{clip}(x,0,\mu)=0, \text{ if } x<0; \ { x}, \text{ if } 0\leq x\leq \mu; \ \mu, \text{ if } x\geq\mu$,  and $X_j$ and $W_{ij}$ denote the outputs of the neurons in the preceding layer and the weights connecting the two layers. The  gradients of $\mu$ are estimated using gradient descent during the backward computations of the DNN.  

On the other hand, the computation dynamics of a SNN are typically represented by the popular Leaky-Integrate-and-Fire (LIF) model \cite{leefin2020}, where a neuron transmits binary spike trains (except the input layer for direct encoding) over multiple time steps ($1$ denotes the presence of a spike). To account for the temporal dimension of the inputs, each input has an internal state called a membrane potential, $U_i(t)$ which captures the integration of the incoming (pre-neuron) spikes (denoted as $S_j(t)$) modulated by weights $W_{ij}$ and leaks with a fixed time constant. Each neuron emits an output spike whenever $U_i(t)$ crosses a spiking threshold $V^{th}$ after which $U_i(t)$ is reduced by $V^{th}$. This behavior of the membrane potential and output can be expressed as
\begin{align}
U_i^{temp}(t)&=\lambda U_i(t-1)+\sum_j W_{ij}{S_j(t)} \\ 
S_i(t)&=
\begin{cases}
    V^{th}, & \text{if } U_i^{temp}(t)>V^{th}\label{eq:lif_output}\\
    0, & \text{otherwise} \\
\end{cases} \\
U_i(t) &= U_i^{temp}(t)-S_i(t)
\label{eq:IF_out_spike}
\end{align}
where $\lambda$ denotes the leak term. When $\lambda=1$, the SNN model is termed Integrate-and-Fire (IF).

\subsection{DNN-to-SNN Conversion}\label{subsec:ann_snn_conversion}

Previous research has demonstrated that SNNs can be converted from DNNs with negligible accuracy drop by approximating the activation value of ReLU neurons with the firing rate of IF neurons using a threshold balancing technique that copies the weights from the source DNN to the target SNN \cite{dsnn_conversion1,dsnn_conversion_cont,dsnn_conversion_ijcnn,dsnn_conversion_abhronilfin}. Since this technique uses the standard backpropagation algorithm for DNN training, and thus involves only a single forward pass to process a single input, the training procedure is simpler than the approximate gradient techniques used to train SNNs from scratch. However, the key disadvantage of DNN-to-SNN conversion is that it yields SNNs with much higher latency compared to other techniques. Some previous research \cite{Han_2020_CVPR,li2021free} proposed to down-scale the threshold term to train low-latency SNNs, but the scaling factor was either a hyperparameter or obtained via linear grid-search, and the latency needed for convergence still remained large (${>}64$). 

\begin{figure}[t!]
\begin{center}
\includegraphics[width=0.5\textwidth]{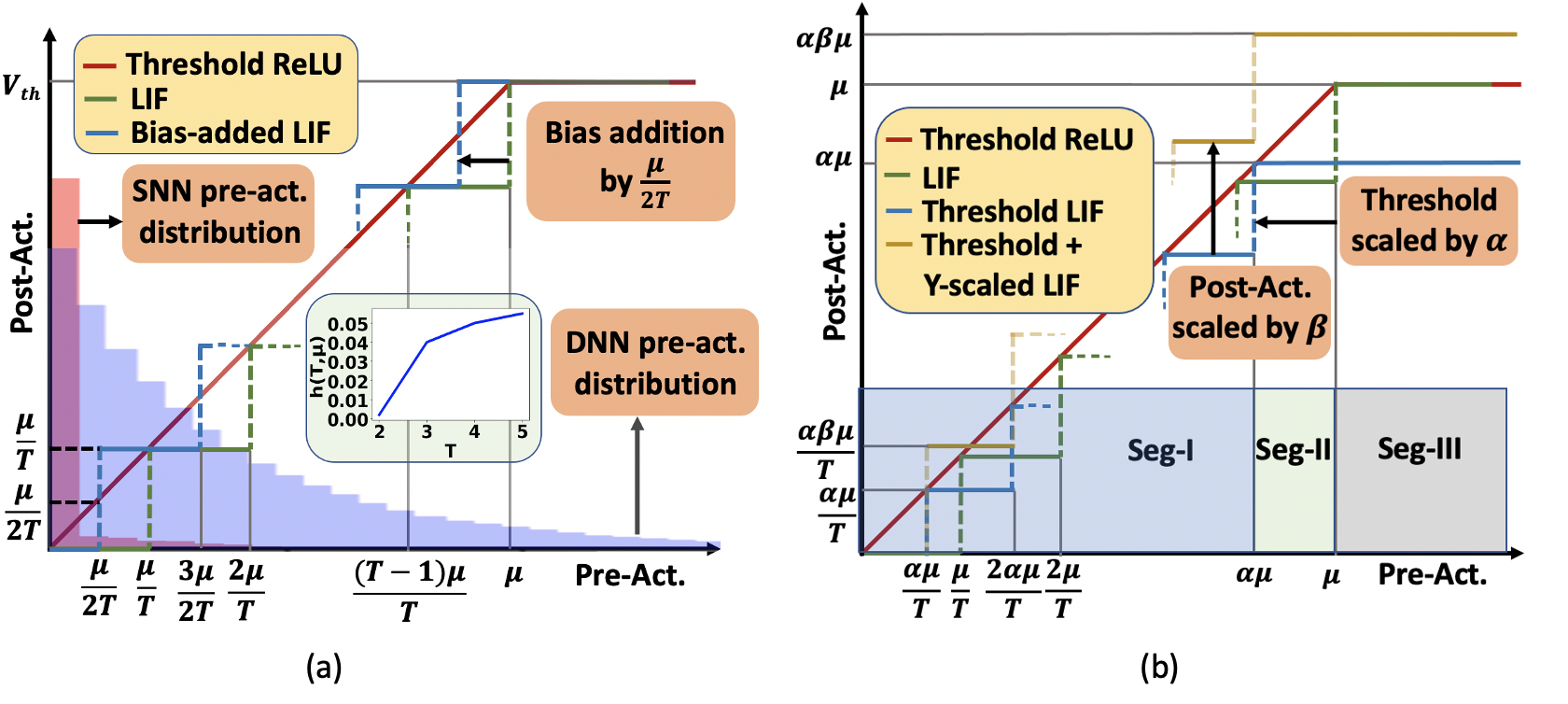}
\end{center}
\vspace{-0.5cm}
\caption{(a) Comparison between DNN (threshold ReLU) and SNN (both original and bias-added) activation functions, the distribution of DNN and SNN ($T=2$) pre-activation values and variation of $h(T,\mu)$ (see Eq. \ref{eq:activation_diff_2}) with $T(\leq5)$ for the $2^{nd}$ layer of VGG-$16$ architecture on CIFAR-$10$, and (b) Proposed scaling of the threshold and output of the SNN post-activation values.}
\label{fig:dnn_snn_activation}
\vspace{-2mm}
\end{figure}

To further reduce the conversion error,  \cite{deng2021optimal} minimized the difference between the DNN and SNN post-activation values for each layer. To do this, the activation function of the IF SNN must first be derived \cite{deng2021optimal, li2021free}. We assume that the initial membrane potential of a layer $l$ ($\textbf{U}_l(0)$) is $0$. Moreover, we let $\Bar{\textbf{S}_l}$ be the average SNN output of layer 
$l$. Then, $\Bar{\textbf{S}_l}=\frac{1}{T}\sum_{i=1}^{T}S_l(i)$ where $S_l(i)$ is the discrete output at the $i^{th}$ time step, and $T$ is the total number of time steps,
\begin{align}\label{eq:snn_activation_diff}
    \Bar{\textbf{S}_l} &= \frac{V^{th}}{T}\text{clip}\left(\floor{\frac{T}{V^{th}}\textbf{W}_l\Bar{\textbf{S}}_{l-1}},0,T\right),
\end{align}
where $V^{th}$ and $W_l$ denote the layer threshold and weight matrix respectively. Eq \ref{eq:snn_activation_diff} is illustrated in Fig. \ref{fig:dnn_snn_activation}(a) by the piecewise staircase function of the SNN activation.

Reference \cite{deng2021optimal} also proved that the average difference in the post-activation values can be reduced by adding a bias term $\delta$ to shift the SNN activation curve to the left by $\delta{=}{V^{th}}/{2T}$, as shown in Fig. \ref{fig:dnn_snn_activation}(a), assuming both the DNN ($d$) and SNN ($s$) pre-activation values are \textit{uniformly and identically distributed}. To further reduce the difference, \cite{deng2021optimal} added a non-trainable threshold equal to the maximum DNN pre-activation $(d_{max})$ value to the ReLU activation function in each layer and equated it with the SNN spiking threshold, which 
ensures zero difference between the DNN and SNN post-activation values when the DNN pre-activation values exceed $d_{max}$. However, $d_{max}$ is an outlier, and ${>}{99\%}$ of the pre-activation values lie between $[0,\frac{d_{max}}{3}]$. Hence, we propose to use the ReLU activation with a trainable threshold for each layer (denoted as $\mu$, where $\mu{<}d_{max}$ for all layers) as discussed in Section \ref{sec:background} and shown in Fig. \ref{fig:dnn_snn_activation}(a). This trainable threshold, as will be described below, also helps reduce the average difference for non-uniform DNN pre-activation distributions.

\section{Proposed Training Framework}\label{sec:ann_snn_conversion_failure}

In this section, 
we analytically and empirically show that the SOTA conversion strategies, along with our proposed modification described above, fail to obtain the SOTA SNN test accuracy for smaller time steps. 
We then propose a novel conversion algorithm that scales the SNN threshold and post-activation values to reduce the conversion error for small $T$. 

\subsection{Why Does Conversion Fail for Ultra Low Latencies?}\label{subsec:reduce_conversion_error}
\begin{figure}[t!]
\begin{center}
\includegraphics[width=0.3\textwidth]{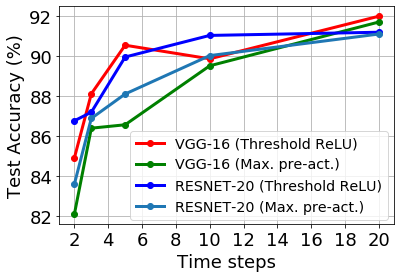}
\end{center}
\vspace{-5mm}
\caption{Effect of the number of SNN time steps on the test accuracy of VGG and ResNet architectures on CIFAR-$10$ with DNN-to-SNN conversion based on both threshold ReLU and the maximum pre-activation value used in \cite{deng2021optimal}.}
\label{fig:acc_vs_time-steps_curve}
\vspace{-5mm}
\end{figure}

Even though we can minimise the difference between the DNN and SNN post-activation values with bias addition and thresholding, in practice, the SNNs obtained are still not as accurate as their iso-architecture DNN counterparts when $T$ decreases substantially. We empirically show this trend for VGG and ResNet architectures on the CIFAR-$10$ dataset in Fig. \ref{fig:acc_vs_time-steps_curve}. 
\emph{This is due to the flawed baseline assumption that 
the DNN and SNN pre-activation are uniformly distributed}. Both the distributions are rather skewed (i.e., most of the values are close to $0$), as illustrated in Fig. \ref{fig:dnn_snn_activation}(a). 

To analytically see this, let us assume the DNN and SNN pre-activation probability density functions are $f_D(d)$ and $f_S(s)$ and post-activation values are denoted as $d'$ and $s'$, respectively. Assuming $V^{th}{=}{\mu}$, derived from DNN training, the \textit{expected} difference in the post-activation values $\Delta=E(d')-E(s')$ for a particular layer and $T$ can be written as 
\begin{align}\label{eq:activation_diff_1}
     \Delta&\approx \int\limits_{0}^{\mu}\left( d'f_D(d)\partial d{-}s'f_S(s)\partial s \right)=\int\limits_{0}^{\mu}\left( df_D(d)\partial d{-}s'f_S(s)\partial s \right) \notag \\
     &= K(\mu)\mu{-}\left(\sum_{i=1}^{T-1}i(\frac{\mu}{T})g_i(T,\mu)\right){-}{\mu}{\int\limits_{T'}^{\mu}{f_S(s)\partial s}},
\end{align}
where the first approximation is due to the fact that greater than ${99.9\%}$ of both $d$ and $s$ are less than $\mu$. The subsequent equality is because $d'=d$ when $d \leq \mu$. The last equality is based on the introduction of $g_i(T,\mu)=\int\limits_{(i-\frac{1}{2})\frac{\mu}{T}}^{(i+\frac{1}{2})\frac{\mu}{T}}f_S(s)\partial s$ which captures the bias shift of $\frac{\mu}{2T}$, and $T'={(T-\frac{1}{2}\mu)}/{T}$ and the observation that the term $\int\limits_{0}^{\mu}df_D(d)\partial d$ lies between its upper and lower integral limits, and thus can be re-written as $K(\mu)\mu$, where $K(\mu)$ lies in the range $[0,1]$. The exact value of $K(\mu)$ depends on the distribution $f_D(d)$.

Assuming $h(T,\mu){=}\left(\sum_{i=1}^{T-1}(\frac{i}{T})g_i(T,\mu)\right){+}{\int\limits_{T'}^{\mu}{f_S(s)\partial s}}$, Eq. \ref{eq:activation_diff_1} can be then written as
\begin{align}\label{eq:activation_diff_2}
     \Delta&\approx\mu(K(\mu)-h(T,\mu)).
\end{align}

When $f_D(d)$ and $f_S(s)$ are uniformly distributed in the range $[0,\mu]$, they must equal ${1}/{\mu}$. This implies that $\int\limits_{0}^{\mu}df_D(d)\partial d=\frac{\mu}{2}$ and, consequently, $K(\mu)=\frac{1}{2}$. Moreover, $g_i(T,\mu)=\frac{1}{T} \; \forall  i{\in}[1,(T{-}1)]$, and hence the first term of $h(T,\mu)$, $\sum_{i=1}^{T-1}(\frac{i}{T})g_i(T,\mu)=\frac{T-1}{2T}$, whereas the second term, ${\int\limits_{T'}^{\mu}{f_S(s)\partial s}}$, equals  $\frac{1}{2T}$. Hence, similar to $K(\mu)$, $h(T,\mu)=\frac{1}{2}$. Thus, Eq. \ref{eq:activation_diff_2} evaluates to $0$ which implies the error can be completely eliminated, as also concluded in \cite{deng2021optimal}.

However, when the distributions are skewed, we observe that while $K(\mu)$ is independent of $T$, $h(T,\mu)$ decreases significantly as we reduce $T$ below around 5, 
as shown in the insert in Fig. \ref{fig:dnn_snn_activation}(a). Intuitively, for small $T$, most of the probability density of $s$ lies to the left of the first staircase starting at $s{=}\frac{\mu}{2T}$,  
due to its sharply decreasing nature.
Consequently, the remaining area under the curve captured in $h(T,\mu)$ becomes negligible, reducing the number of output spikes significantly. 

Hence, for ultra-low SNN latencies, the error $\Delta$ per layer remains significant and accumulates over the network. 

This analysis explains the accuracy gap that is observed between original DNNs and their converted SOTA SNNs for $T \leq 5$, as exemplified in Fig. \ref{fig:acc_vs_time-steps_curve}. Moreover, training with a non-trainable threshold \cite{deng2021optimal}, can be modeled by 
replacing $\mu$ with $d_{max} \geq \mu$ in Eq. \ref{eq:activation_diff_2}.
This further increases $\Delta$, as observed from the increased accuracy degradation shown in Fig. \ref{fig:acc_vs_time-steps_curve}.

\vspace{-2mm}
\subsection{Conversion \& Fine-tuning for Ultra Low-Latency SNNs}\label{subsec:proposed_conversion_methodology}
While Eq. \ref{eq:activation_diff_2} suggests that we can tune $\mu$ to compensate for low $T$, this introduces other errors. In particular, if we replace $\mu$ with a down-scaled version\footnote{Up-scaling $\mu$ further reduces the output spike count and increases the error.} $\alpha \mu$, with $\alpha\in(0,1)$,
the SNN activation curve will shift left, as shown in Fig. \ref{fig:dnn_snn_activation}(b), and there will be an additional difference between $d'$ and $s'$ that stems from the 
values of $d$ and $s$ in the range $(\alpha \mu, \mu)$ as follows
\begin{align}
\Delta_{\alpha} &\approx\alpha\mu(K(\alpha\mu){-}h(T,\mu))+ \int\limits_{\alpha\mu}^{\mu}\left(d'f_D(d)\partial d-s'f_S(s)\right)\partial s \notag \\ 
&=\alpha\mu(K(\alpha\mu)-h(T,\mu)){+}{\int\limits_{\alpha \mu}^{\mu}df_D(d)\partial d}-\alpha\mu\int\limits_{\alpha\mu}^{\mu}f_S(s)\partial s \notag 
\end{align}
To mitigate this additional error term, we propose to also optimize the step size of the SNN activation function in the y-direction by modifying the IF model from Eq. \ref{eq:lif_output},
\begin{align}\label{eq:modified_lif}
S_i^{\beta}(t)&=
\begin{cases}
    \beta V^{th}, & \text{if } U_i^{temp}(t)>V^{th}\\
    0, & \text{otherwise}, \\
\end{cases}
\end{align}
which introduces another scaling factor $\beta$ illustrated in Fig. \ref{fig:dnn_snn_activation}(b). Moreover, we remove the bias term since it complicates the parameter space exploration and poses difficulty in training the SNNs after conversion, changing $h(T,\mu)$ to $h'(T,\mu)=\sum_{i=1}^{T-1}(\frac{i}{T})g_{(i-1/2)}(T)$. 
This results in a new difference function 
\begin{align}
\Delta_{\alpha\beta}{\approx}\alpha\mu(K(\alpha\mu){-}\beta h'(T,\mu)){+}{\int\limits_{\alpha\mu}^{\mu}df_D(d)\partial d} {-}\alpha\beta\mu\int\limits_{\alpha\mu}^{\mu}f_S(s)\partial s \notag \label{eq:activation_diff_with_scaling_beta}
\end{align}
Thus, our task reduces to finding the $\alpha$ and $\beta$ that minimises $\Delta_{\alpha\beta}$ for a given low $T$. 

Since it is difficult to analytically compute $\Delta_{\alpha \beta}$ to guide SNN conversion, we empirically estimate it by discretizing $d$ into percentiles $P[j]$ $\forall{j}\in\{0,1,..,M\}$, where $M$ is the largest integer satisfying $P[M] \leq \mu$, using the activations of a particular layer of 
the trained DNN. 
In particular, for each $\alpha=\frac{P[j]}{\mu}$, we vary 
$\beta$ between $0$ and $2$ with a step size of $0.01$, as shown in Algorithm \ref{alg:x+y_scale_compute}.
This percentile-based approach for $\alpha$ is better than a linear search because it enables a finer-grained analysis in the range of $d$ with higher likelihood. We find the ($\alpha, \beta$) pair that yields the lowest $\Delta_{\alpha\beta}$ for each DNN layer. 

For DNN-to-SNN conversion, we copy the SNN weights from a pretrained DNN with trainable threshold $\mu$, set each layer threshold as $\alpha\mu$, and produce an output $\beta V^{th}$ whenever the membrane potential crosses the threshold. Although we incur an overhead of two additional parameters per SNN layer, the parameter increase is negligible compared to the total number of weights. Moreover, as the outputs for each time step are either $0$ or $\beta V^{th}$, we can absorb the scaling factor into the weight values, avoiding the need for explicit multiplication. After conversion, we apply SGL in the SNN domain where we jointly fine-tune the threshold, leak, and weights \cite{rathi2020dietsnn}. 
To approximate the gradient of ReLU, we compute the surrogate gradient as $\frac{\partial s'}{\partial s}\approx  1, \text{ if } 0\leq s \leq 2\alpha\mu$, and $0$ otherwise, which is used to estimate the gradients of the trainable parameters \cite{rathi2020dietsnn}.

\begin{algorithm}[t]
\small
\SetAlgoLined
\DontPrintSemicolon
\textbf{Input}: Activations \textbf{$A$}, Total time steps {T}, ReLU threshold $\mu$ \\
\KwData{$ i = 0,1,...M$, percentiles $P[i]=i^{th} \text{percentile of } A$, where $M$ is the largest integer satisfying $P[M]\leq\mu$, initial scaling factors $\alpha^i=1$ and $\beta^i=1$} 
\textbf{Output}: Final scaling factors $\alpha^f$ and $\beta^f$ \\
\SetKwFunction{FMain}{ComputeLoss}
    \SetKwProg{Fn}{Function}{:}{}
    \Fn{\FMain{$P$,$\mu$,$\alpha$,$\beta$,$T$}}{
        $ loss \leftarrow 0 $    \\

        {\ForEach{$ p \in P$}
        {\For{$j \leftarrow 0$ \KwTo $(T-1)$}
        {\If{$\frac{j\alpha\mu}{T}\leq p\leq \frac{(j+1)\alpha\mu}{T}$ } 
            {$ loss \leftarrow loss + (p-\frac{j\alpha\beta\mu}{T})$   \#Seg-I in Fig. \ref{fig:dnn_snn_activation}(b)
} }
        }
            {
            {\If{$\alpha\mu<p\leq\mu$}
            {$loss \leftarrow loss+ (p-\alpha\beta\mu)$ \#Seg-II in Fig. \ref{fig:dnn_snn_activation}(b) 
            }
            {\uElseIf{$p>\mu$}
            {$loss \leftarrow loss+ \mu(1-\alpha\beta)$ \#Seg-III in Fig. \ref{fig:dnn_snn_activation}(b) 
            }}
            }}

        }
        
        \textbf{return} $loss$ 
}
\textbf{End Function}. \\
\SetKwFunction{FMain}{\textbf{FindScalingFactors}}
    \SetKwProg{Fn}{Function}{:}{}
    \Fn{\FMain{$P$,$\mu$,$T$}}{
    $preloss=$ \texttt{ComputeLoss} $(P, \mu, \alpha^i, \beta^i, T)$ \\
    \ForEach{$ p \in P $}{
    {\For{$ j \leftarrow 0$ \KwTo $2$ (step size of $0.01$)}
    {$loss  = $ \texttt{ComputeLoss}$(P, \mu, \frac{p}{\mu}, j, T)$ \\
    {\If{$|loss|<|preloss|$}
    {$\alpha^f = \frac{p}{\mu}$, $\beta^f = j$,  $preloss = loss$   \\
    
    }
    }
    }
    }
    }
    \textbf{return} ${\alpha^f}, {\beta^f}$
    }
\textbf{End Function}. \\

\caption{Detailed algorithm for finding layer-wise scaling factors for SNN threshold \& post-activations}
\label{alg:x+y_scale_compute}
\end{algorithm}

\section{Experimental Results}\label{sec:expt_results}

\subsection{Experimental Setup}

Since we omit the bias term during DNN-to-SNN conversion described in Section \ref{subsec:proposed_conversion_methodology}, we avoid Batch Normalization, and instead use Dropout as the regularizer for both ANN and SNN training. Although prior works \cite{rathi2020dietsnn, dsnn_conversion_abhronilfin, rathi2020iclr} claim that max pooling incurs information loss for binary-spike-based activation layers, we use max pooling because it improves the accuracy of both the baseline DNN and converted SNN.
Moreover, max pooling layers produce binary spikes at the output, and ensures that the SNN only requires AC operations for all the hidden layers \cite{fang2021incorporating}, thereby improving energy efficiency.

We performed the baseline DNN training for $300$ epochs with an initial learning rate (LR) of $0.01$ that decays by a factor of $0.1$ at every $60$, $80$, and $90\%$ of the total number of epochs. Initialized with the layer thresholds and post-activation values, we performed the SNN training with direct input encoding for $200$ epochs for CIFAR-$10$ and $300$ epochs for CIFAR-$100$. 
We used a starting LR of $0.0001$ which decays similar to that in DNN training. All experiments are performed on a Nvidia $2080$ Ti GPU with $11$GB memory.

\begin{table}[!t]
\begin{center}
\scriptsize\addtolength{\tabcolsep}{-1pt}
\begin{tabular}{|c|c|c|c|c|}
\hline
 & Number & a. & b. Accuracy ($\%$) & c. Accuracy ($\%$)  \\
 Architecture & of & DNN ($\%$) & with DNN-to-SNN & after SNN \\
 &  time steps & accuracy     &  conversion    & training    \\ 
\hline
\hline
 \multicolumn{5}{|c|}{Dataset : CIFAR-10} \\
\hline
\hline
 VGG-11  & 2  & 90.76 & 65.82 & 89.39  \\
        & 3 & 91.10 & 78.76 & 89.79  \\
\hline
 VGG-16 & 2   & 93.26  & 69.58  &  91.79  \\
        & 3 & 93.26  & 85.06 & 91.93   \\
\hline
 ResNet-20 &2 & 93.07 & 61.96  & 90.00 \\
        & 3 & 93.07 & 73.57 &  90.06 \\
\hline
\hline
 \multicolumn{5}{|c|}{Dataset : CIFAR-100} \\
\hline
\hline
 VGG-16  & 2 & 68.45 & 19.57 & 64.19  \\
        & 3 & 68.45 & 36.84 & 63.92 \\
\hline
 ResNet-20  & 2 & 63.88 & 19.85 & 57.81 \\
        & 3 & 63.88 & 31.43 & 59.29 \\
\hline

\end{tabular}
\end{center}
\caption{Model performances with proposed training framework after a) DNN training, b) DNN-to-SNN conversion \& c) SNN training.}
\label{tab:dnn_snn_results}
\vspace{-7mm}
\end{table}

\subsection{Classification Accuracy \& Latency}
We evaluated the performance of these networks on multiple VGG and ResNet architectures, namely VGG-$11$, and VGG-$16$, and Resnet-$20$ for CIFAR-$10$, VGG-$16$ and Resnet-$20$ for CIFAR-$100$. 
We report the (a) baseline DNN accuracy, (b) SNN accuracy with our proposed DNN-to-SNN conversion, and (c) SNN accuracy with conversion, followed by SGL, for $2$ and $3$ time steps. Note that the models reported in (b) are far from SOTA, but act as a good initialization for SGL. 

Table \ref{tab:snn_comparison} provides a comparison of the performances of models generated through our training framework with SOTA deep SNNs. On CIFAR-$10$, our approach outperforms the SOTA VGG-based SNN \cite{rathi2020dietsnn} with $2.5\times$ fewer time steps and negligible drop in test accuracy. To the best of our knowledge, our results represent the first successful training and inference of CIFAR-$100$ on an SNN with only $2$ time steps, yielding a $2.5{-}8\times$ reduction in latency compared to others. 

\textit{Ablation Study:} The threshold scaling heuristics proposed in \cite{li2021free,Han_2020_CVPR}, coupled with SGL, lead to a statistical test accuracy of ${\sim}10\%$ and ${\sim}1\%$ on CIFAR-$10$ and CIFAR-$100$ respectively, with both $2$ and $3$ time steps. 
Also, our scaling technique alone (without SGL) requires $12$ steps, while the SOTA conversion approach \cite{deng2021optimal} needs $16$ steps to obtain similar test accuracy.
\vspace{-1mm}

\section{Simulation Time \& Memory Requirements}\label{sec:train_memory_analysis}

\begin{figure*}[t!]
\begin{center}
\includegraphics[width=\textwidth]{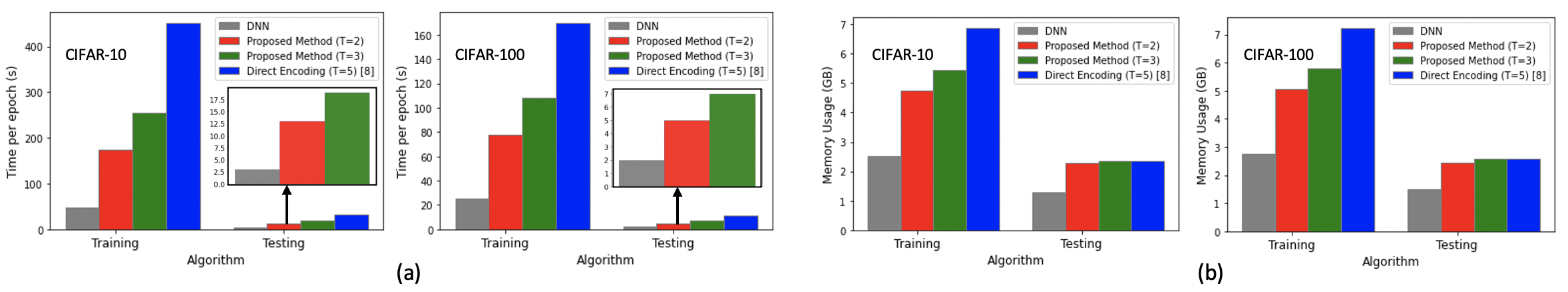}
\end{center}
\vspace{-0.6cm}
\caption{Comparison between our proposed hybrid training technique for $2$ and $3$ time steps, baseline direct encoded training for $5$ time steps \cite{rathi2020dietsnn} based on (a) simulation time per epoch, and (b) memory consumption, for VGG-$16$ architecture over CIFAR-$10$ and CIFAR-$100$ datasets}
\label{fig:time+memory_ananlysis}
\vspace{-5mm}
\end{figure*}

Because SNNs require iteration over multiple time steps and storage of the membrane potentials for each neuron, their simulation time and memory requirements can be substantially higher than their DNN counterparts. However, reducing their latency can bridge this gap significantly, as shown in Figure \ref{fig:time+memory_ananlysis}. On average, our low-latency, 2-time-step SNNs represent a $2.38\times$ and $2.33\times$ reduction in training and inference time per epoch respectively, compared to the hybrid training approach \cite{rathi2020dietsnn} which represents the SOTA in latency, with iso-batch conditions. Also, our proposal uses $1.44\times$ lower GPU memory compared to \cite{rathi2020dietsnn} during training, while the inference memory usage remains almost identical.  

\begin{table}
\begin{center}
\scriptsize\addtolength{\tabcolsep}{-1pt}
\begin{tabular}{|c|c|c|c|c|}
\hline
Authors &  Training & Architecture & Accuracy & Time  \\
 & type &  & ($\%$) & steps \\
\hline
\hline
 \multicolumn{5}{|c|}{Dataset : CIFAR-10} \\
\hline
\hline
Wu et al. & Surrogate & 5 CONV, & 90.53 & 12 \\
(2019) \cite{wu2019direct} & gradient & 2 linear &  &  \\
\hline
Rathi et al. & Hybrid & VGG-$16$ & 92.70 & 5  \\
(2020) \cite{rathi2020dietsnn} & training  &   &  & \\
\hline
Kundu et al. & Hybrid & VGG-$16$ & \textbf{92.74}  & 10 \\
(2021) \cite{kundu2021lowlatency} & training & &   &  \\
\hline
Deng et al. & DNN-to-SNN & VGG-16 & 92.29 & 16 \\
(2021) \cite{deng2021optimal}& conversion & & & \\
\hline
This work & Hybrid Training & VGG-16 & {91.79} & \textbf{2} \\
\hline
\hline
 \multicolumn{5}{|c|}{Dataset : CIFAR-100} \\
\hline
\hline
Kundu et al. & Hybrid & VGG-$16$ & 65.34 & 10 \\
(2021) \cite{kundu2021lowlatency}& training & CNN & &\\
\hline
Deng et al. & DNN-to-SNN & VGG-$16$ &  \textbf{65.94} & 16 \\
(2021) \cite{deng2021optimal}& conversion & & &  \\
\hline
This work & Hybrid Training & VGG16 & {64.19} &  \textbf{2} \\
\hline
\end{tabular}
\end{center}
\caption{Performance comparison of the proposed training framework with state-of-the-art deep SNNs on CIFAR-10 and CIFAR-100.}
\label{tab:snn_comparison}
\vspace{-6mm}
\end{table}

\begin{figure*}[t!]
\begin{center}
\includegraphics[width=\textwidth]{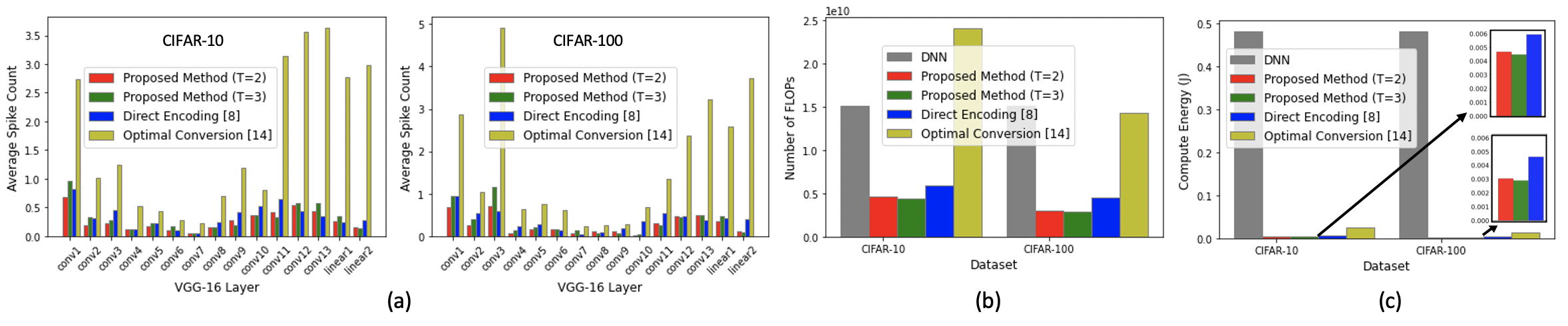}
\end{center}
\vspace{-0.6cm}
\caption{Comparison between our proposed hybrid training technique for $2$ and $3$ time steps, baseline direct encoded training for $5$ time steps \cite{rathi2020dietsnn}, and the optimal DNN-to-SNN conversion technique \cite{deng2021optimal} for $16$ time steps, based on (a) average spike count, (b) total number of FLOPs, and (c) compute energy, for VGG-$16$ architecture over CIFAR-$10$ and CIFAR-$100$ datasets. An iso-architecture DNN is also included for comparison of FLOP count and compute energy.}
\label{fig:energy_test_analysis}
\vspace{-0.3cm}
\end{figure*}

\section{Energy Consumption During Inference}\label{sec:energy_analysis}
\subsection{Spiking Activity}

As suggested in \cite{kundu_2021,datta2021training}, the average spiking activity of an SNN layer $l$ can be used as a measure of the compute energy of the model during inference. 
This is computed as the ratio of the total number of spikes in $T$ steps over all the neurons of the layer $l$ to the total number of neurons in that layer. 
Fig. \ref{fig:energy_test_analysis}(a) shows the per-image average number of spikes for each layer with our proposed algorithm (using both $2$ and $3$ time steps), the hybrid training algorithm by \cite{rathi2020dietsnn} (with $5$ steps), and the SOTA conversion algorithm \cite{deng2021optimal} which requires $16$ time steps, 
while classifying CIFAR-$10$ and CIFAR-$100$ using VGG-$16$. 
On average, our approach yields $1.53\times$ and $4.22\times$ reduction in spike count compared to \cite{rathi2020dietsnn} and \cite{deng2021optimal}, respectively.

\subsection{Floating Point Operations (FLOPs) \& Compute Energy}

We use FLOP count to capture the energy efficiency of our SNNs, since each emitted spike indicates which weights need to be accumulated at the post-synaptic neurons and results in a fixed number of AC operations. This, coupled with the MAC operations required for direct encoding in the first layer (also used in \cite{rathi2020dietsnn,deng2021optimal}), dominates the total number of FLOPs. 
For DNNs, FLOPs are dominated by the MAC operations in all the convolutional and linear layers. Assuming $E_{MAC}$ and $E_{AC}$ denote the MAC and AC energy respectively, the inference compute energy of the baseline DNN model can be computed as $\sum_{l=2}^{L}FL^l_D\cdot E_{AC}$, whereas that of the SNN model as $FL^1_S\cdot E_{MAC}+\sum_{l=2}^{L}FL^l_S\cdot E_{AC}$, where $FL^l_D$ and $FL^l_S$ are the FLOPs count in the $l^{th}$ layer of DNN and SNN respectively.

Fig. \ref{fig:energy_test_analysis}(b) and (c) illustrate the FLOP counts and compute energy consumption for our baseline DNN and SNN models of VGG16 while classifying CIFAR-datasets, along with the SOTA comparisons \cite{rathi2020dietsnn, deng2021optimal}. As we can see, the number of FLOPs for our low-latency SNN is smaller than that for an iso-architecture DNN and the SNNs obtained from the prior works. 
Moreover, ACs consume significantly less energy than MACs both on GPU as well as neuromorphic hardware. To estimate the compute energy, we assume a $45$ nm CMOS process at $0.9$ V, where $E_{AC}=0.1$ pJ, while $E_{MAC}=3.2$ pJ $(3.1$ for multiplication and $0.1$ for addition) \cite{horowitz20141} for 32-bit integer representation. Then, for CIFAR-$10$, our proposed SNN consumes $103.5\times$ lower compute energy compared to its DNN counterpart and $1.27\times$ and $5.18\times$ lower energy than \cite{rathi2020dietsnn} and \cite{deng2021optimal} respectively. For CIFAR-$100$, the improvements are $159.2\times$ over the baseline DNN, $1.52\times$ over the 5-step hybrid SNN, and $4.72\times$ over the 16-step optimally converted SNN. 

On custom neuromorphic architectures, such as TrueNorth \cite{Merolla2014AMS}, and SpiNNaker \cite{spinnaker}, the total energy is estimated as $FLOPs{*}{E_{compute}}{+}{T}{*}{E_{static}}$ \cite{park_2020}, where the parameters $(E_{compute},E_{static})$ can be normalized to $(0.4, 0.6)$ and $(0.64, 0.36)$ for TrueNorth and SpiNNaker, respectively \cite{park_2020}. Since the total FLOPs for VGG-$16$ $({>}{10^9})$ is several orders of magnitude higher than the SOTA $T$, the total energy of a deep SNN on neuromorphic hardware is compute bound and thus we would see similar energy improvements on them.

\section{Conclusions}\label{sec:conclude}
This paper shows that current DNN-to-SNN algorithms cannot achieve ultra low latencies
because they rely on simplistic assumptions of the DNN and SNN pre-activation distributions. 
The paper then proposes a novel training algorithm, inspired by empirically observed distributions, that can more effectively optimize the SNN thresholds and post-activation values. This approach 
enables training of SNNs with as little as $2$ time steps and without any significant degradation in accuracy for complex image recognition tasks. The resulting SNNs are estimated to consume $159.2\times$ lower energy than iso-architecture DNNs. 


\bibliographystyle{IEEEtran}
\bibliography{IEEEabrv, biblio}

\end{document}